\documentclass[letterpaper, 10 pt, conference]{ieeeconf}  

\IEEEoverridecommandlockouts                              

\usepackage{color}
\usepackage{amsmath}
\usepackage{lipsum}
\usepackage{gensymb}
\usepackage{tabularx}
\usepackage{multicol}
\usepackage{graphicx}
\usepackage{float}
\graphicspath{ {./images/} }
\usepackage{subfig}

\title{\LARGE \bf
An Interactive Indoor Drone Assistant
}

\author{Tino Fuhrman$^{1}$, David Schneider$^{1}$, Felix Altenberg$^{1}$, Tung Nguyen$^{1}$,\\ Simon Blasen$^{1}$, Stefan Constantin$^{1}$, Alex Waibel$^{1,2}$
\thanks{$^{1}$Interactive Systems Lab, Institute for Anthropomatics and Robotics
        Karlsruhe Institute of Technology, 76137 Karlsruhe, Germany
        {\tt\small tino.fuhrmann@student.kit.edu}}%
\thanks{$^{2}$School of Computer Science, Carnegie Mellon, 
        5000 Forbes Ave, Pittsburgh, PA 15213
        {\tt\small waibel@andrew.cmu.edu}}%
}

\begin{document}

\maketitle
\thispagestyle{empty}
\pagestyle{empty}

\begin{abstract}
With the rapid advance of sophisticated control algorithms, the capabilities of drones to stabilise, fly and manoeuvre autonomously have dramatically improved, enabling us to pay greater attention to entire missions and the interaction of a drone with humans and with its environment during the course of such a mission. In this paper, we present an indoor office drone assistant that is tasked to run errands and carry out simple tasks at our laboratory, while given instructions from and interacting with humans in the space. To accomplish its mission, the system has to be able to understand verbal instructions from humans, and perform subject to constraints from control and hardware limitations, uncertain localisation information, unpredictable and uncertain obstacles and environmental factors. We combine and evaluate the dialogue, navigation, flight control, depth perception and collision avoidance components. We discuss performance and limitations of our assistant at the component as well as the mission level.  A 78\% mission success rate was obtained over the course of 27 missions.

\end{abstract}

\section{INTRODUCTION}

Drone technology and drone control has recently advanced rapidly to the point that
consumer drones are already commonplace, displaying impressive features and capabilities.
Particularly, advanced sensors and improved control algorithms have made flying drones much
simpler, more performant and made a variety of drone applications (aerial surveys, mapping,
aerial movies and even selfie-drones) possible.  As the flight capabilities of these devices improve,
interacting with drones and providing them with their mission interactively, become increasingly
important challenges.  How can we point, gesture and speak with a drone, so it “knows” what it
should  do?  How can it take instructions and corrections mid-flight?  Can it gradually become a
personal assistant, along with humanoid robots, and other AI “agents”?

In this paper, we present work towards a speaking assistant drone.  The drone is
intended to fly missions at our laboratory to look for people, objects, items, and show people around.
Also, to be safe in an indoor environment, it must be small and quiet. Much of the computational work can be run on servers, provided that the drone has fast
communication links and sensors.

We present our first prototype, which accepts directions by voice dialogue as to who to visit in a
laboratory.  It must deal with a realistic, changing environment, varying obstacles and lighting
conditions, and it should be able to accept correction and new missions in-flight. 

In the following, we present our system design, and report our experiments flying the
drone.  We discuss and evaluate the performance of the system components, as well as
of overall mission success.

\section{RELATED WORK}
Autonomous flying robots are a fast growing research area. While outdoor navigation of Unmanned Aerial Vehicles (UAVs) is often GPS-based, indoor flights face many challenges due to the lack of an external positioning system. SLAM (simultaneous localisation and mapping) is a popular approach and has shown good results \cite{ monoslam, indoor4}. However SLAM-based systems usually require sophisticated hardware, for example laser scanners and depth cameras. For a lightweight platform such as the one we discuss in this paper, visual methods based machine learning techniques may be a viable alternative. In order to save weight on the drone itself, many of the more computationally intense calculations can be done by a separate computer,  provided the communication with the drone is fast enough. Such methods have already shown promising results \cite{indoor7}. Deep neural networks (DNNs) have also been applied to directly navigate quadcopters in unknown indoor environment \cite{indoor1, indoor2, interpercept}.

Some approaches have also attempted to use neural networks to directly calculate a collision probability from a 2D image \cite{dronet,pulp} rather than first generating a 3D map of the environment.

With the rapid advancement of Automatic Speech Recognition (ASR) technologies, there are initiatives to make use of these developments in order to interact with flying robots. Recent works to develop voice activated ground control stations for aerial vehicles have yielded promising results \cite{vgcs1, vgcs2}. As to the best of our knowledge, state-of-the-art systems for voice-controlled drones re so far able to recognise a limited number of commands and translate these directly into fixed controls signals \cite{vdrone1, vdrone2, vdrone3}.

\section{APPROACH}\label{sec:approach}

Our approach is based on the notion of missions. A mission consists of input
parameters and success conditions called goals. A mission is considered to be
completed successfully if all goals are achieved. In the
case of the system presented in this paper, the input parameter is a verbal
request to fly to a certain destination (room or person) in an office environment. The goal
of the mission is to reach the target without any manual intervention and collision with static or dynamic obstacles. 

\begin{figure} [h]
    \centering
    \includegraphics[width=0.48\textwidth]{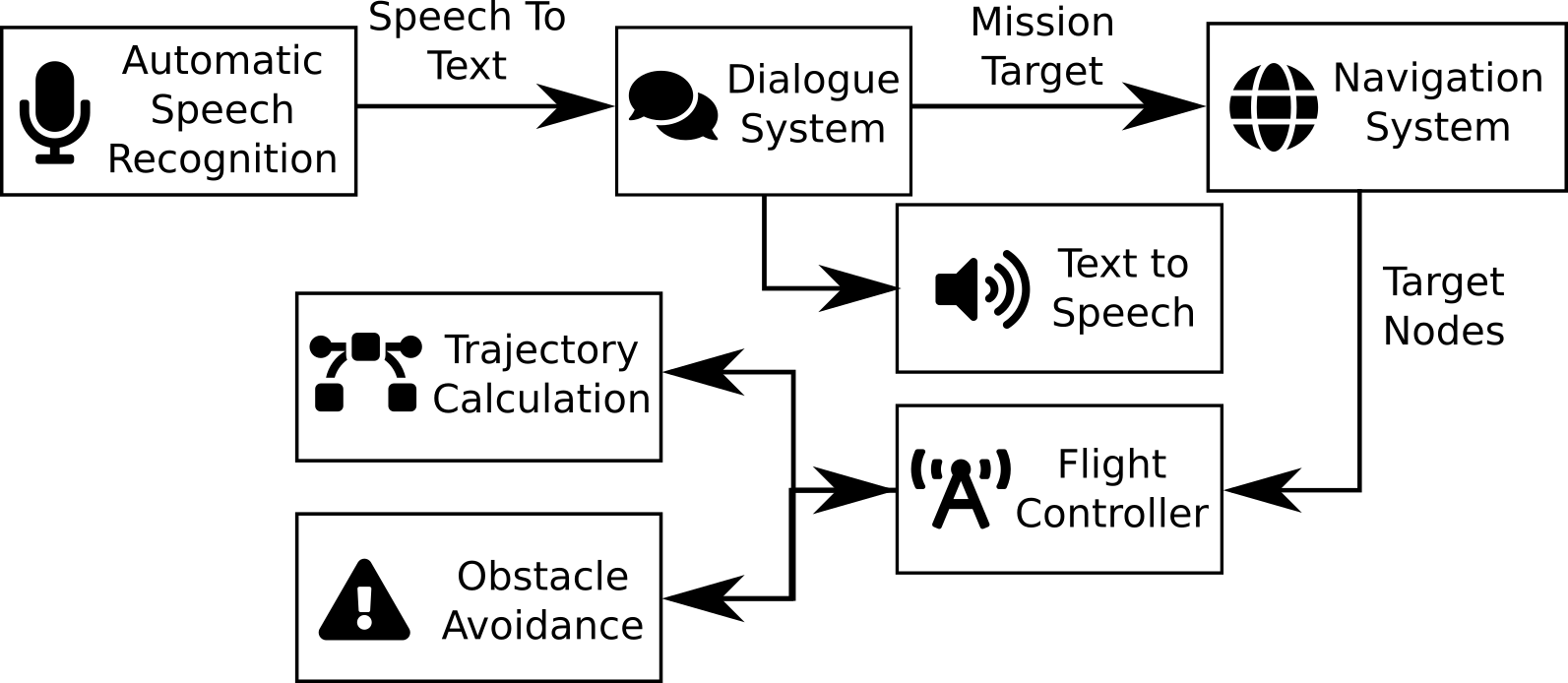}
    \begin{flushright}
    \footnotesize{
          Icons~courtesy~of~Fonticons,~Inc.
    }
    \end{flushright}
    \caption{System Overview}
    \label{fig:subsystems}
\end{figure}

This description already suggests that the system must be decomposed into
multiple subsystems. Figure \ref{fig:subsystems} shows the decomposition of our system into these subsystems. First, the ASR subsystem 
transcribes the verbal request and forwards it to the dialogue system. The
dialogue system then identifies the target office and passes it to the
navigation system. From here, the flight controller uses the obstacle
avoidance component and the navigation system to follow a collision-free
trajectory to the target of the mission. During this process, the coordination
of the flight control components is crucial for achieving a high degree of
safety and reliability. 

In the following sections, each of the individual components is discussed in more details.
\subsection{Platform and Customisations}

\begin{figure}[h]
    \centering
    \includegraphics[width=0.4\textwidth]{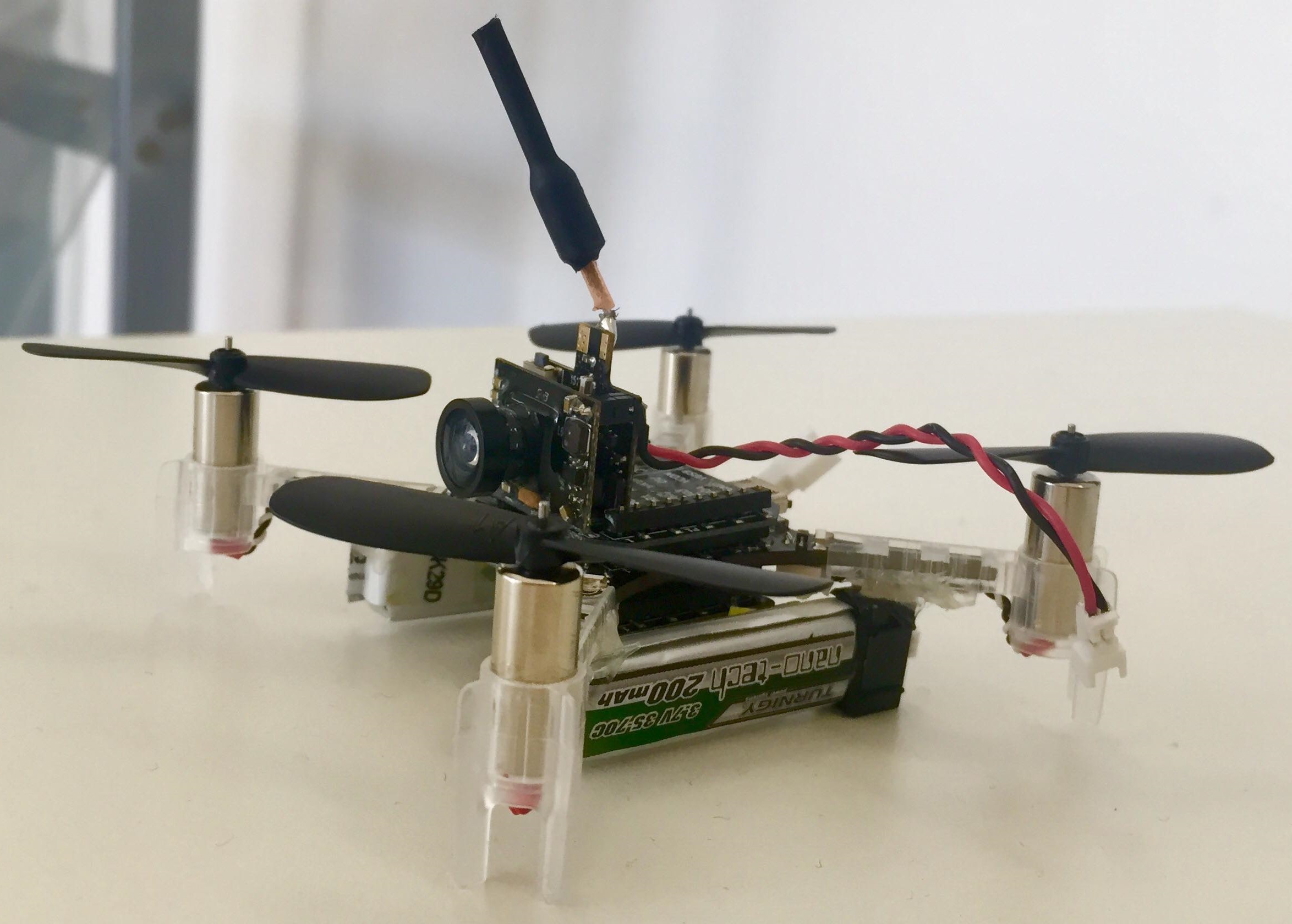}
    \caption{One of our \textit{Crazyflie} drones}
    \label{fig:setup}
\end{figure}

For our experiments we use the \textit{Crazyflie 2.0} quadcopter by  \textit{Bitcraze} \cite{giernacki2017crazyflie} with some slight modifications. What sensors could be used in this project were limited by the \textit{Crazyflie}'s small size and weight capacity. Our setup uses the \textit{Flow Deck}, an optical flow sensor provided by \textit{Bitcraze} \cite{flowdeck} that can measure movement in relation to the ground. A small camera was added in order to
capture images to be used for depth perception. The original battery was replaced by two
larger batteries in order to accommodate the additional power requirement and the
motors were upgraded in order to be able to more easily carry the increased
total weight of approximately 43 grams. A picture of one of our drones can be seen in Figure \ref{fig:setup}.

The drone is controlled by a computer which receives data from the drone's
on-board sensors and sends control commands back. This communication is done
using the \textit{Crazyradio PA} \cite{bitcrazeWiki}, a 2.4
GHz radio antenna provided by \textit{Bitcraze} specifically for use with the
\textit{Crazyflie}. Additionally, the video signal from the camera is
transmitted separately through a 5.8 GHz radio system to provide a fast "First Person View" image directly from the drone. This provides a
live video feed from the drone that can then be evaluated. The video signal is then further passed on to a server with a high-end GPU which evaluates the pictures and provides a real time depth map that can then be used for obstacle avoidance.

\subsection{Dialogue Based Mission Control}
Our dialogue system consists of three components: First, the automatic speech
recognition (ASR) component, which transcribes the recorded audio into text.
Second, the dialogue component, which extracts the semantic meaning from the
transcription, manages the dialogue, and generates natural language output. Third,
the text to speech (TTS) component, which synthesises the textual natural
language output from the dialogue component to spoken language.

For the TTS system, we use an external component, because this component needs
not to be adapted to the domain of drone mission control.

As our ASR component we use the Janus speech recognition toolkit \cite{LevinLWGGKW2000, nguyen2019improving}. We added our dialogue system training dataset to the language model of
the ASR component and the words of the training dataset to the vocabulary of the ASR
component to improve the recognition.

The dialogue component is based on an attention-based encoder-decoder model
\cite{BahdanauCB15} and is trained end-to-end. The input is the dialogue history
and the current utterance of the user. We use byte pair encoding
\cite{SennrichHB2016}. The output is either an API call, e.\,g. \textit{FlyTo
ROOM 232}, or a natural language output, e.\,g. \textit{Sorry, I don't know where that is}.

\subsection{Localisation and Trajectory Planning}

For positioning and navigation we currently use an implementation based on the
drone's local coordinate frame, which is supported by our object avoidance
approach. The optical flow sensor provides us with the drones position in 3D in reference to a fixed
starting point. For each deck the sensor was calibrated to give as accurate readings of distance as possible.

\begin{figure}[h]
    \centering
    \includegraphics[width=0.48\textwidth,]{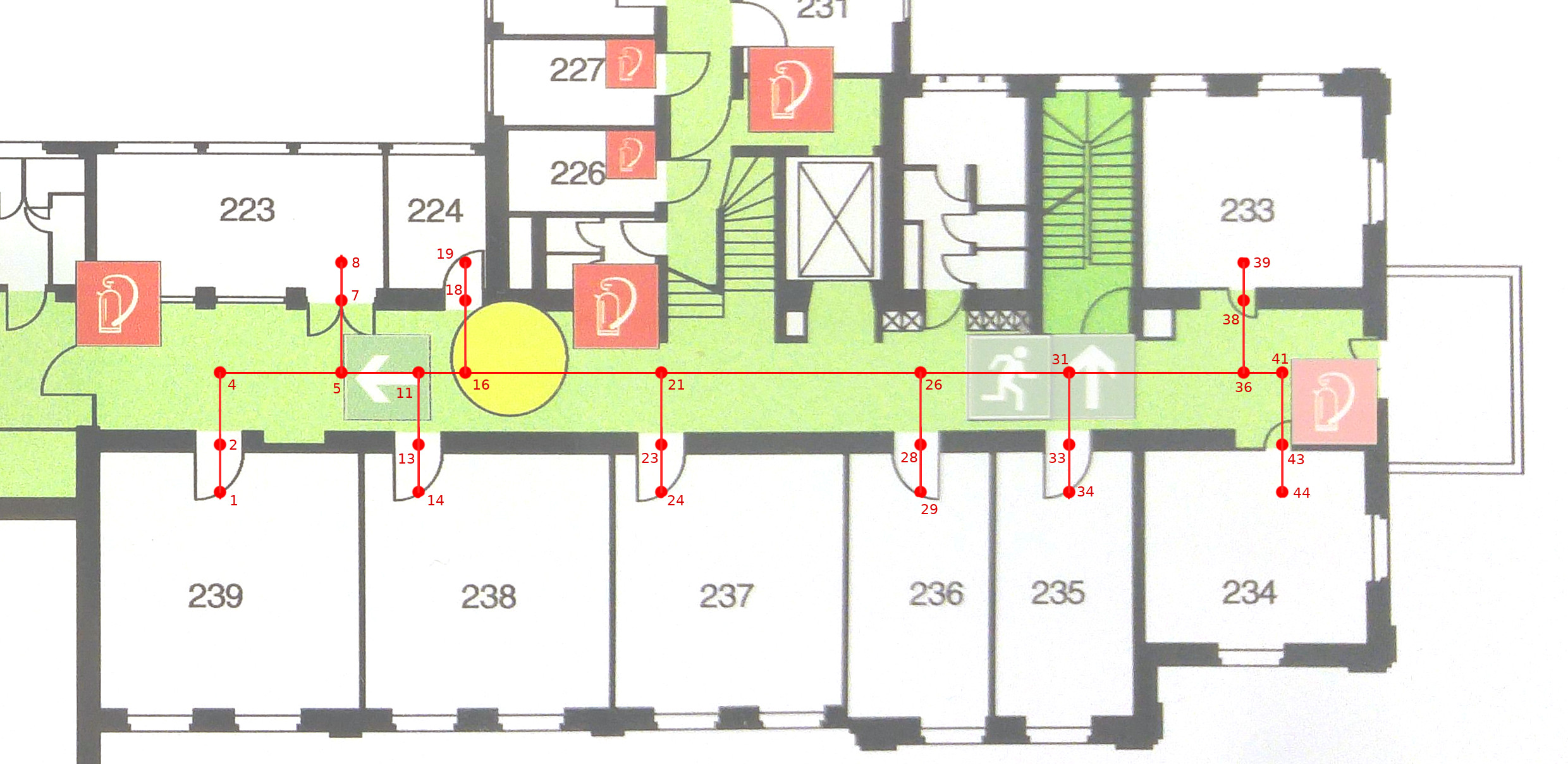}
    \caption{Diagram of our corridor with the nodes the system uses to navigate shown in red}
    \label{fig:nodemap}
\end{figure}

Using the drone's current position in relation to a fixed reference frame, basic
navigation is performed by evaluating a map, see Figure \ref{fig:nodemap}, consisting of few important
connected nodes which are positioned along hallways or door openings and enable
the drone to calculate a basic trajectory through passageways instead of trying
to approach the target directly. The points are augmented with information about
the room number they belong to support navigation requests such as ``{}Fly to
room 235'' or ``Fly to Stefan's office''. Dijkstra's algorithm
\cite{dijkstra1959note} based on euclidean distance is applied to find the
shortest sequence of points between the current location of the drone and the target.
When a navigation request is received, the system first identifies the closest 
navigation point, then identifies the target node which corresponds to the navigation request. It then
calculates the shortest trajectory between start and target and uses the
resulting points as a sequence of direct targets pending any corrections from the collision avoidance system.

Simply marking a node as reached when the drone is less than a certain distance
away, causes the drone to often circle the target point due to drift, noise in
the \textit{Flow Deck}'s position estimation or interference from the obstacle avoidance
subsystem. To counteract this behaviour and to enhance the smoothness of the flight
trajectory, a line perpendicular to the planned flight direction (the line
defined by the most recently reached node and the next node) with a certain
distance to the next node is calculated. The drone is then considered to have 
reached the vertex, if it has crossed this line.

\subsection{Object Avoidance}

As previously mentioned, the sensors that were available for this project were  limited by the weight the \textit{CrazyFlie} is able to carry. For this reason, 
it was not possible to add additional sensors such as LIDAR to obtain 
more accurate distance measurements. Instead, we opted for using a mono RGB camera 
with a 160\degree~field of view as the sole sensor for obstacle avoidance. 

First, a depth map is obtained from the camera feed using a state-of-the-art
neural network approach. Several depth prediction neural networks
\cite{obstacle:monodepth, obstacle:fcrn} were evaluated in an indoor
environment. During the
evaluation of the different approaches, it became clear that only the approach
published in \cite{obstacle:fcrn} generalised well to our specific requirements. In particular, no retraining was required to obtain usable depth maps in office environments with our camera. The selected method uses a single RGB image as input
for a fully convolutional neural network to
produce a $160 \times 128$ pixel depth map. Samples of RGB images and the
corresponding depth maps produced by the neural network can be seen in Figure
\ref{fig:obstacle_avoidance}. As can be observed in the middle column of this figure, the network is more accurate for obstacles close to the camera than for obstacles farther away.

The next step of the obstacle avoidance subsystem is determining in which
directions the drone can fly. To determine if an obstacle is in front of
the drone, the depth map is divided into $k = 9$ vertical stripes. For each stripe
$s_i$, the share of pixels with a value greater than $\epsilon$ is
calculated. $\epsilon$ was determined empirically by evaluating multiple depth maps. 
To reduce noise, a majority vote using the last
$n = 4$ depth maps is used to determine if a pixel is above the threshold
$\epsilon$.

\begin{figure}[h]
    \centering
    \includegraphics[width=0.4\textwidth,]{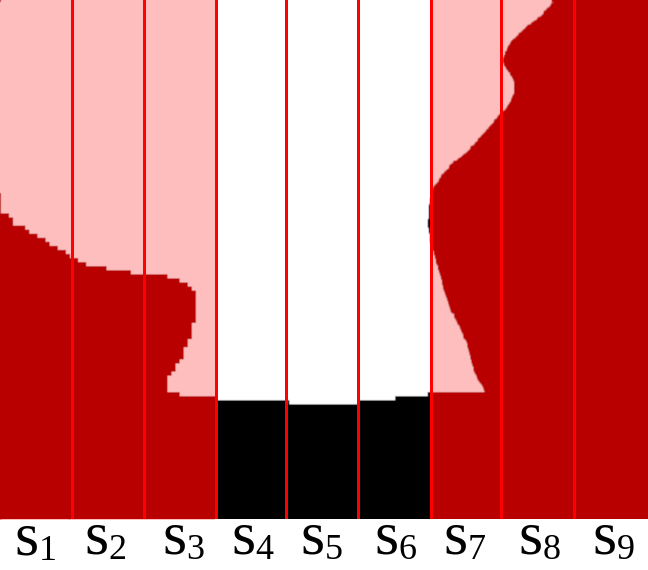}
    \caption{$s_4, s_5, s_6$ are deemed \textit{obstacle-free}. This includes the middle stripe $s_5$ and so the drone decides to continue flying forwards.}
    \label{fig:depth_map_example}
\end{figure}

If a stripe $s_i$ has a share of more than 70\%, it is called
\textit{obstacle-free}. If this is the case for the stripe in the middle of the
depth map, it is assumed that the drone can continue to fly straight.
Otherwise, if any of the two stripes to the left or right of the middle stripe
are \textit{obstacle-free}, the drone is instructed to fly to the left or
right respectively to avoid the obstacle in front of it. Furthermore, the speed
of the drone in \textit{x} and \textit{y} direction is adapted based on
$s_{\frac{k}{2}}$ to reduce the probability of colliding with an obstacle due to
excessive speed or inaccurate depth predictions. An example of this process can be seen in Fig. \ref{fig:depth_map_example}. This simple decision making process is
sufficient to avoid most static and dynamic obstacles in an office environment.

\section{Experiments}

For the purposes of testing, each previously mentioned component of the system was evaluated separately.

\subsection{Dialogue Based Mission Control}

\begin{figure*}[!htb]
\subfloat{  
      \includegraphics[height=4.4cm]{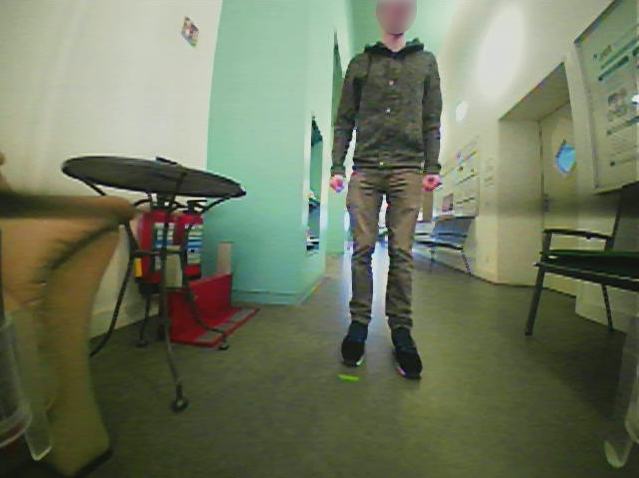}
}
\subfloat{
          \includegraphics[height=4.4cm]{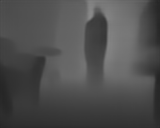}
}
\subfloat{
      \includegraphics[height=4.4cm]{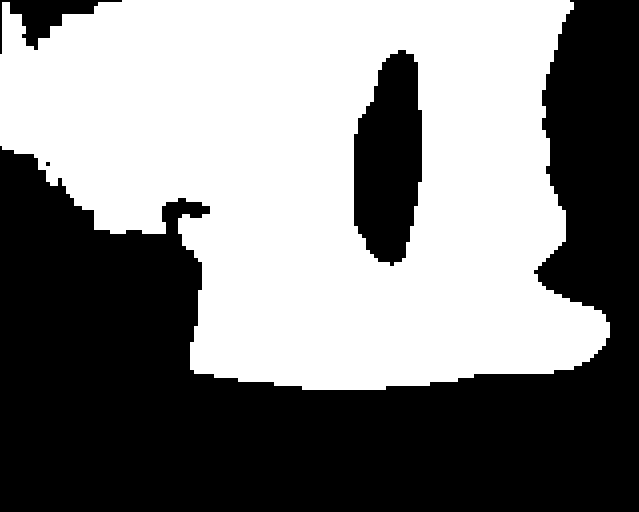}
}
\setcounter{subfigure}{0}
\subfloat[Camera Image]{
  \includegraphics[height=4.4cm]{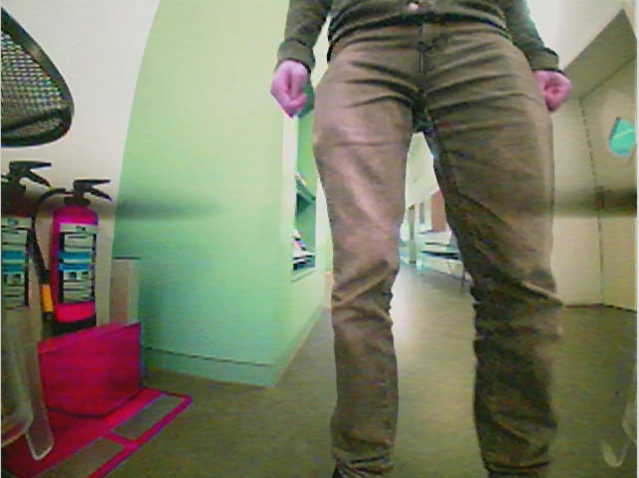}
}
\subfloat[Depth Map]{
      \includegraphics[height=4.4cm]{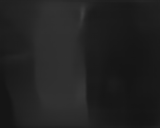}
}
\subfloat[Obstacle-Free Areas]{
      \includegraphics[height=4.4cm]{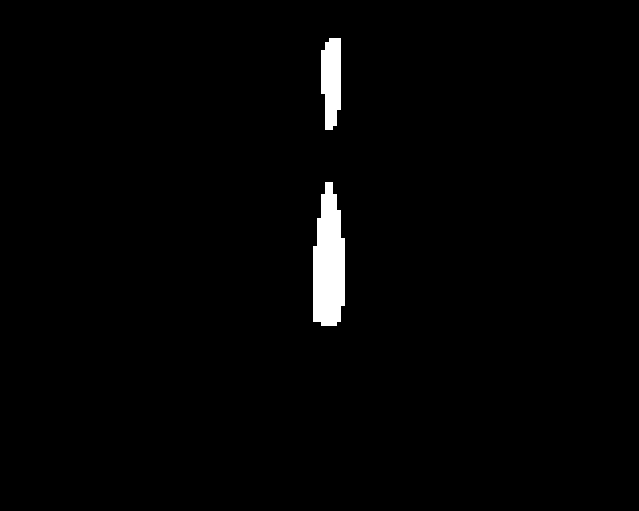}
}
\caption{Overview of the different images used by the Obstacle Avoidance System}\label{fig:obstacle_avoidance}
\end{figure*}

For the evaluation of the dialogue based mission control system, we asked three non-native English speakers to order the drone to fly to a room or visit a person in their room. 
The volunteers were provided with a list of names of persons (some of which were not known by the dialogue system) and a list of rooms.
They were then told to use simple sentences to direct the drone to a room or person of their choosing. 
We provided no examples before the experiment in order to realistically capture how different persons interact with the drone.
Table \ref{tab:sample_sentences} shows some examples of the sentences used by our volunteers.
Table \ref{tab:dialog_results} shows the accuracy of the target room detected by the dialogue system grouped by volunteer and goal type (either fly to room or person).

\begin{table}[h]
\begin{tabularx}{\linewidth}{|X|p{1cm}|}
\hline
\textbf{Command} & \textbf{Test Subject} \\ \hline
Now I want you to go to room two hundred thirty eight? & 1\\ \hline
If you would be so kind, could you please go to Peter?  & 1 \\ \hline
Can you fly Professor Waibel? & 3 \\ \hline
Can you fly to room two two three? & 2 \\ \hline
Perfect. And after that could you please go to Stefan Constantin? & 1  \\ \hline
\end{tabularx}
\caption{example sentences used by the volunteers}\label{tab:sample_sentences}
\end{table}

\begin{table}[h]
\begin{tabularx}{\linewidth}{|X|X|X|X|X|}
\hline
\textbf{Test Subject} & \textbf{Number of Tests}& \textbf{Goal: Room} & \textbf{Goal: Person} & \textbf{Total}  \\ \hline
    1  &14    &   75\% & 50\% & 57\%  \\ \hline
    2  &13    &    66\% & 30\% & 38\% \\ \hline
    3  &11    &    100\% & 16 \% & 45\% \\ \hline
\end{tabularx}
\caption{accuracy of the dialogue based mission control grouped by volunteer and goal type}\label{tab:dialog_results}
\end{table}

Across all volunteers, it can be observed that the system recognised commands asking the drone to fly to a specific room (e.g. room 223) with a higher accuracy than commands ordering the drone to go to a room of a specific person.
Due to the various ways in which names can be pronounced, they are more complicated to recognise for the ASR component. 
Furthermore, the pronunciation of names varies between different languages which in particular is problematic for non-native speakers. Note that the first volunteer's accent was closest to that of a native English speaker, while the second and third volunteer had heavier accents.

These difficulties can cause the ASR component to transcribe names or complete sentences (in the case of a heavy accent) incorrectly and make it impossible for the dialogue system to recognise the intended target room.
In all experiments in which the drone could not correctly infer the user's goal, the dialogue system correctly identified that it can not determine the goal of the mission and asked the user to restate their order.

\subsection{Depth Perception and Collision Avoidance}
\begin{figure}[ht]
    \centering
    \includegraphics[width=0.35\textwidth, angle=270]{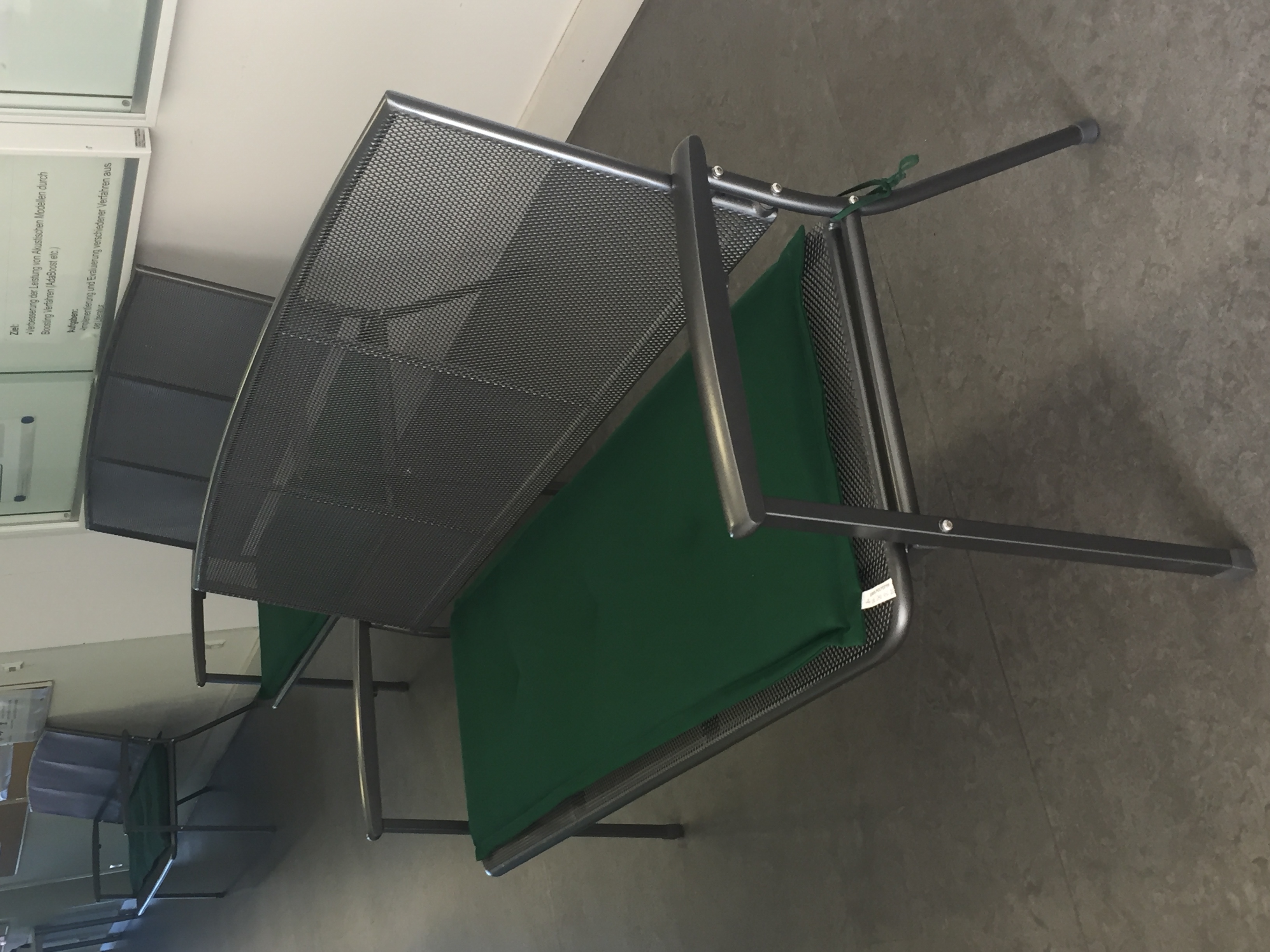}
    \caption{The bench used for testing the collision avoidance system}
    \label{fig:Bench}
\end{figure}

In order to test the drone's collision avoidance system, we confronted the drone with three different kinds of obstacle. The first was a closed doorway as we expected this to be relatively easy for the drone to detect. The second was a person blocking the flight path of the drone. This would be an interesting test to see if maybe the drone could find a way around the person and continue on its way. The third object we used to test the drone was a metal bench, see Figure \ref{fig:Bench}. We knew from preliminary testing that the system struggles to see objects that are very narrow, such as the armrests or the legs of a bench or desk, and objects that are partially translucent. The bench would be a good test of this as it incorporates both those elements.

The drone performed very well in the first test.
Out of 8 attempts to fly through the closed door the drone never collided with the door, always stopping well before. The average distance the the drone stopped before the door was 93 $\pm$ 21cm. While there does seem to be a high variance in the distance that the drone stops at, it never came closer than 63cm to the door which can be considered a safe distance.

Testing with a person in the path of the drone was similarly successful, once again stopping in time on 8 out of 8 attempts, and in 6 attempts the drone managed to find a way around the person to keep going. One of these attempts can be seen in Figure \ref{fig:obstacle_avoidance}.

As predicted the test using the bench were not as successful. In order to prevent any damage being done to the drone the experiment was stopped after 4 attempts after the drone flew into the bench on all 4 of said attempts.

\subsection{Mission Completion}
In order to test the rate of mission completion the drone was sent to various rooms along the corridor using textual input to specify the target room. The path to the room was always left unobstructed. Furthermore, only rooms were chosen that would keep the drone within range of the control computer. The dialogue system was excluded from this test due to the large variance in performance between different users. However, note that the combined success rate could be calculated as the joint probability of the success rate of the flight control and the dialogue system. 

\begin{table}[h]
\begin{tabular}{|l|l|ll}
\hline
\textbf{Mission}                                                       & \textbf{Success} & \multicolumn{1}{l|}{\textbf{\begin{tabular}[c]{@{}l@{}}Time \\ in s\\ ($\pm 1s$) \end{tabular}}} & \multicolumn{1}{l|}{\textbf{Comments}}                                                                               \\ \hline
Fly 223-\textgreater{}238                                              & Yes              & \multicolumn{1}{l|}{76}                                                            & \multicolumn{1}{l|}{}                                                                                                \\ \hline
Fly 223-\textgreater{}238                                              & Yes              & \multicolumn{1}{l|}{47}                                                            & \multicolumn{1}{l|}{}                                                                                                \\ \hline
Fly 223-\textgreater{}238                                              & Yes              & \multicolumn{1}{l|}{56}                                                            & \multicolumn{1}{l|}{}                                                                                                \\ \hline
Fly 237-\textgreater{}238                                              & Yes              & \multicolumn{1}{l|}{85}                                                            & \multicolumn{1}{l|}{}                                                                                                \\ \hline
Fly 237-\textgreater{}238                                              & No               & \multicolumn{1}{l|}{-}                                                             & \multicolumn{1}{l|}{\begin{tabular}[c]{@{}l@{}}Got stuck due to thinking a\\ bookshelf was in its way\end{tabular}} \\ \hline
Fly 238-\textgreater{}223                                              & No               & \multicolumn{1}{l|}{-}                                                             & \multicolumn{1}{l|}{Drone twisted during takeoff}                                                                    \\ \hline
Fly 238-\textgreater{}223                                              & Yes              & \multicolumn{1}{l|}{118}                                                           & \multicolumn{1}{l|}{\begin{tabular}[c]{@{}l@{}}Got stuck for a while due\\ to strong sunlight\end{tabular}}          \\ \hline
Fly 238-\textgreater{}223                                              & Yes              & \multicolumn{1}{l|}{83}                                                            & \multicolumn{1}{l|}{}                                                                                                \\ \hline
Fly 238-\textgreater{}223                                              & No               & \multicolumn{1}{l|}{-}                                                             & \multicolumn{1}{l|}{Drone twisted during takeoff}                                                                    \\ \hline
Fly 238-\textgreater{}223                                              & Yes              & \multicolumn{1}{l|}{121}                                                           & \multicolumn{1}{l|}{\begin{tabular}[c]{@{}l@{}}Got stuck for a while on a\\ desk's shadow\end{tabular}}              \\ \hline
Fly 238-\textgreater{}223                                              & Yes              & \multicolumn{1}{l|}{44}                                                            & \multicolumn{1}{l|}{}                                                                                                \\ \hline
Fly 238-\textgreater{}223                                              & Yes              & \multicolumn{1}{l|}{40}                                                            & \multicolumn{1}{l|}{}                                                                                                \\ \hline
Fly 238-\textgreater{}223                                              & Yes              & \multicolumn{1}{l|}{39}                                                            & \multicolumn{1}{l|}{}                                                                                                \\ \hline
Fly 238-\textgreater{}237                                              & No               & \multicolumn{1}{l|}{-}                                                             & \multicolumn{1}{l|}{Drone twisted during takeoff}                                                                    \\ \hline
Fly 238-\textgreater{}237                                              & Yes              & \multicolumn{1}{l|}{64}                                                            & \multicolumn{1}{l|}{}                                                                                                \\ \hline
Fly 238-\textgreater{}237                                              & Yes              & \multicolumn{1}{l|}{79}                                                            & \multicolumn{1}{l|}{}                                                                                                \\ \hline
Fly 238-\textgreater{}237                                              & Yes              & \multicolumn{1}{l|}{78}                                                            & \multicolumn{1}{l|}{}                                                                                                \\ \hline
Fly 238-\textgreater{}237                                              & Yes              & \multicolumn{1}{l|}{84}                                                            & \multicolumn{1}{l|}{}                                                                                                \\ \hline
Fly 238-\textgreater{}239                                              & Yes              & \multicolumn{1}{l|}{206}                                                           & \multicolumn{1}{l|}{\begin{tabular}[c]{@{}l@{}}Got stuck due to bad depth\\ perception\end{tabular}}                 \\ \hline
Fly 238-\textgreater{}239                                              & Yes              & \multicolumn{1}{l|}{85}                                                            & \multicolumn{1}{l|}{}                                                                                                \\ \hline
Fly 238-\textgreater{}239                                              & Yes              & \multicolumn{1}{l|}{59}                                                            & \multicolumn{1}{l|}{}                                                                                                \\ \hline
Fly 238-\textgreater{}239                                              & Yes              & \multicolumn{1}{l|}{72}                                                            & \multicolumn{1}{l|}{}                                                                                                \\ \hline
Fly 238-\textgreater{}239                                              & Yes              & \multicolumn{1}{l|}{40}                                                            & \multicolumn{1}{l|}{}                                                                                                \\ \hline
Fly 238-\textgreater{}239                                              & Yes              & \multicolumn{1}{l|}{54}                                                            & \multicolumn{1}{l|}{}                                                                                                \\ \hline
Fly 239-\textgreater{}238                                              & No               & \multicolumn{1}{l|}{-}                                                             & \multicolumn{1}{l|}{Lost Connection}                                                                 \\ \hline
Fly 239-\textgreater{}238                                              & No               & \multicolumn{1}{l|}{-}                                                             & \multicolumn{1}{l|}{\begin{tabular}[c]{@{}l@{}}Strong sunlight glare affecting \\ the depth perception\end{tabular}} \\ \hline
Fly 239-\textgreater{}238                                              & Yes              & \multicolumn{1}{l|}{60}                                                            & \multicolumn{1}{l|}{}                                                                                                \\ \hline
\textbf{\begin{tabular}[c]{@{}l@{}}Total Success\\ Rate:\end{tabular}} & 77.78\%          &                                                                                    &                                                                                                                      \\ \cline{1-2}
\end{tabular}
\caption{Table showing the Mission results sorted by Mission}\label{tab:missionResults}
\end{table}

We can see from the results in Table \ref{tab:missionResults} that the overall success rate is quite good at 77.78\% over 27 missions. One of the more common causes of mission failure is that the drone turns slightly during takeoff, often due to the 4 propellers not all turning on at the same time. This causes a problem due to the fact that the drone reports its current position relative to where it started. A change in the initial yaw angle of the drone to report its current position but with an offset angle relative to its starting position, which in turns causes the entire internal map the computer uses to control the drone to be shifted by that angle.

Most of the other causes of failure are due to problems with the depth perception. The neural network extracting depth information from flat images is not always completely accurate. For example it sometimes thinks an object's shadow on the ground is an obstacle that can not be flown through. In rare cases it simply fails to extract enough information from the image due to low contrast, for example if there are not many objects in view or the colours are all very similar. Another common problem is that strong sunlight falling into the camera lens can cause a strong glare effect where most of the picture the camera sends is simply white light. The depth perception system simply views this as an impassable solid object.

It should be noted that usually these problems do not cause the mission to fail completely. Most of the time it simply causes the drone to pause or fly much more slowly until eventually the depth perception system corrects itself. This also explains the sometimes vastly different mission completion times. Sometimes simply moving from one room to another with slightly different lighting conditions can cause the drone to slow down until the camera has adjusted.

\addtolength{\textheight}{-0.2cm}   

\section{Conclusions and Future Work}\label{sec:future_work}

As this is our first prototype there is plenty of room for future improvement, not only on each of the individual components but also the system as a whole.

The results we have seen from the dialogue system were quite disappointing, with recognising the person's intent in, at best, 57\% of the cases. The main problem was that our ASR system had problems recognising the names of the persons and aborts the recognition too early.
In future work, we want to use an improved ASR system.
Furthermore, in order to allow a wider variety of natural language without increasing the size of the training dataset, we also want to use a multi-task approach \cite{ConstantinNW19} in the future.
That means, the drone dataset will be trained alongside an out-of-domain dataset.

The results from the collision detection test were very promising. It managed to stop the drone before colliding into people or large objects and only struggled with very thin or translucent furniture. In order to address this problem in the future we are looking at creating a more precise map of the environment in real time. As previously mentioned, we currently use a pre-recorded 2D map that the system navigates along using the drone's sensor data. The lack of adaptability of this approach also leads to increasingly large errors, especially as the drone gets further away from its starting position. Additionally, this approach cannot handle unknown, static obstacles gracefully and tends to become stuck in these cases. We are looking into using our depth information to create a live 3D map of our environment that we can consistently update. A similar approach was presented in \cite{DBLP:journals/corr/GregorioS17}. 

Reducing these positional errors should also help improve our total mission success rate as this was one of the main causes of mission failure during our tests. The other problem that emerged during our tests was the depth perception system performing badly under very bright or changing light conditions. We plan to also address these issues in the future.

We are also looking at the problem of battery life and battery management. Currently on a full charge the drone can complete 3-4 missions before having to be recharged. In future we not only hope to improve this number but also come up with alternative ways this problem could be mitigated.









\bibliography{bibliography}
\bibliographystyle{IEEEtran.bst}

\end{document}